\pdfoutput=1

\documentclass[11pt]{article}

\usepackage[]{EMNLP2023}

\usepackage{times}
\usepackage{latexsym}

\usepackage[T1]{fontenc}

\usepackage[utf8]{inputenc}

\usepackage{microtype}

\usepackage{inconsolata}

\usepackage{multicol}
\usepackage{enumitem}
\usepackage{linguex}
\usepackage{graphicx}
\usepackage{amssymb}

\title{Can Language Models Be Tricked by Language Illusions? Easier with Syntax, Harder with Semantics}

\author{Yuhan Zhang \\
  Linguistics \\ Harvard University \\
  \texttt{yuz551@g.harvard.edu} \\\And
  Edward Gibson \\
  Brain \& Cognitive Sciences \\ Massachusetts Institute of Technology \\ 
  \texttt{egibson@mit.edu} \\ \And
  Forrest Davis\\
  Computer Science \\ Colgate University \\
  \texttt{fdavis@colgate.edu} \\}

\begin{document}
\maketitle
\begin{abstract}
Language models (LMs) have been argued to overlap substantially with human beings in grammaticality judgment tasks. But when humans systematically make errors in language processing, should we expect LMs to behave like cognitive models of language and mimic human behavior? We answer this question by investigating LMs' more subtle judgments associated with ``language illusions'' – sentences that are vague in meaning, implausible, or ungrammatical but receive unexpectedly high acceptability judgments by humans. We looked at three illusions: the comparative illusion (e.g. ``More people have been to Russia than I have''), the depth-charge illusion (e.g. ``No head injury is too trivial to be ignored''), and the negative polarity item (NPI) illusion (e.g. ``The hunter who no villager believed to be trustworthy will ever shoot a bear''). We found that probabilities represented by LMs were more likely to align with human judgments of being ``tricked'' by the NPI illusion which examines a structural dependency, compared to the comparative and the depth-charge illusions which require sophisticated semantic understanding. No single LM or metric yielded results that are entirely consistent with human behavior. Ultimately, we show that LMs are limited both in their construal as cognitive models of human language processing and in their capacity to recognize nuanced but critical information in complicated language materials.
\end{abstract}

\section{Introduction}

Linguistic evaluations of language models use human language processing data (e.g. human norming data \cite{nair_contextualized_2020,zhang2022representing_new}, acceptability judgments \cite{linzen_assessing_2016,marvin-linzen-2018-targeted}, behavioral or neural measures of language processing \cite{schrimpfNeuralArchitectureLanguage2021,kauf2022event}) as benchmarks to investigate whether LMs possess knowledge of language. This assumes that human-produced data correctly instantiates abstract rules of a language and that humans fully utilize their linguistic knowledge in laboratories and everyday life. However, this assumption is an oversimplification. Humans make consistent errors during language processing \cite{gross_errors_1983}. Under these circumstances, should we expect language models to behave the same as humans? Or should they circumvent human limitations and achieve error-free performance?

Consider, for example, the well-studied case of subject-verb agreement. While we expect an LM of Standard American English to prefer ``the key to the cabinets \textbf{is} on the shelf'' to ``the key to the cabinets \textbf{are} on the shelf'' (as discussed in \citealp{linzen_assessing_2016}), a wealth of psycholinguistic research has systematically documented that humans can ignore errors and accept globally ungrammatical strings (stemming from \citealp{bock1991broken}). Should LMs follow the ideal grammar or mimic human’s (sometimes) errorful behavior?\footnote{For additional critiques of the role of ideal grammatical knowledge in evaluations of LMs, see \citet{pannittoRecurrentBabblingEvaluating2020,weissweiler-etal-2023-construction}.} 

We add to this discussion by investigating three language illusions. Basic examples of each are given in \ref{basic}: the comparative illusion \ref{basicA}, the depth-charge illusion \ref{basicB}, and the negative-polarity item (NPI) illusion \ref{basicC}. All three in \ref{basic} are literally unnatural English sentences, despite the fact that humans often find them surprisingly acceptable. 

\ex.\label{basic}
    \a.\label{basicA} More people have been to Russia than I have.
    \b.\label{basicB} No head injury is too trivial to be ignored.
    \c.\label{basicC} The hunter who no villager believed to be trustworthy will ever shoot a bear.

In this paper, we relied on minimally different strings springing out from the basic illusion sentences that are either (a) considered fully \textbf{acceptable} by human participants, (b) considered fully \textbf{unacceptable} by human participants, or (c) rated \textbf{surprisingly acceptable} by humans (i.e.\ instances of the relevant illusion). We explored whether language models capture the basic contrast between acceptable and unacceptable strings, whether they rate illusion sentences as better than their unacceptable counterparts, and finally, whether models capture nuanced linguistic manipulations that influence human judgments of the illusion material. Further, we compared two ways of measuring models’ preferences, one over the whole sentence (\textit{perplexity}) and another of a privileged position in the sentence (\textit{surprisal}). 

If LMs pattern like human comprehension behavior that involves errors, we expect to derive measures that similarly rate illusion sentences as more acceptable than typical unacceptable sentences. If, on the other hand, LMs align with ideal grammatical judgments, illusion sentences should be rated as unacceptable. Our findings indicate that none of the language models we investigated consistently exhibited illusion effects or demonstrated overall human-like judgment behaviors. Nor do they possess the necessary linguistic knowledge for error-free, literal sentence processing. These findings add more insights into the discussion of LMs' emulation of human behavior and their construal as cognitive models of human language processing.

\section{Related work}

\subsection{LMs' linguistic abilities}

We draw insights from evaluation work relying on acceptability tasks. The construction of minimal pairs has been used to evaluate models for a variety of linguistic processes, including subject-verb agreement 
\citep[e.g.][]{linzen_assessing_2016}, filler-gap dependency \citep[e.g.][]{wilcox_what_2018}, control \citep[e.g.][]{stengel-eskin-van-durme-2022-curious}, and binding \citep[e.g.][]{davis-2022-incremental}. 
This basic template has been expanded into a variety of benchmarks, both for investigations of English 
\citep[e.g.][]{warstadt_blimp_2020},
but also, other languages (e.g. Chinese \citep{song-etal-2022-sling}; Russian \citep{mikhailov-etal-2022-rucola}; 
Japanese \citep{someya-oseki-2023-jblimp}). 
While aggregated results suggest that models overlap with human acceptability judgments in a variety of cases \citep[e.g.][]{hu-etal-2020-systematic}, LMs can behave in distinctly non-human-like ways in capturing the intricacies of grammatical phenomenon \citep[e.g.][]{lee-schuster-2022-language},
the interaction between linguistic processes \citep[e.g.][]{davis-van-schijndel-2020-discourse}, and in generalizing knowledge to infrequent items \citep[e.g.][]{wei_frequency_2021}. 

In our experiments, we are interested in cases where human interpretations and behaviors differ from what is expected given the literal content of the entire string. \textit{Garden path} sentences are a classic example of this basic phenomenon. Strings like ``The horse raced past the barn fell'' are often difficult for humans on first reading because the word \textit{raced} is misparsed as a main verb (e.g. \textit{the horse raced past}) rather than a reduced relative clause (e.g. \textit{the horse that was raced past the barn fell}). LMs have been shown to similarly misprocess these sentences \citep{van_schijndel_single-stage_2021}, though they fall short of capturing the magnitude of the processing cost \citep{arehalli-etal-2022-syntactic}. Here we expand these investigations to language illusions that similarly trigger errorful acceptable judgments in humans while being unnatural and unacceptable. We find that LMs do not pattern like humans in all cases. 

\subsection{Language illusions}
Language illusions refer to ungrammatical, semantically vague, or pragmatically implausible sentences that receive higher than expected acceptability by humans \cite{phillips_grammatical_2011}. 
We study three language illusions in particular: comparative illusion \cite{montalbetti_after_1984} (Section \ref{section:ci}), depth-charge illusion \cite{wason_verbal_1979} (Section \ref{section:dc}), and NPI illusion \cite{xiang_illusory_2009} (Section \ref{section:npi}). Existing human research has found that the illusion effects for both the comparative and the depth-charge illusion are robust and overwhelming but the NPI illusion effect only appears during speeded judgment tasks or word-by-word online paradigms \cite{parker_negative_2016,wellwood_anatomy_2018,paape_quadruplex_2020, orth_negative_2021}.

For human sentence processing, it has been suggested that language illusions provide evidence for rational inference of error-prone strings which integrates heuristics and available context information during processing \cite{ferreira_good-enough_2002,levy_noisy-channel_2008,gibson2013rational, futrell_lossycontext_2020,hahn_resource-rational_2022,zhang2023noisy}. These phenomena raise fundamental questions like what is the role of our grammatical knowledge in comparison to other cognitive resources when it comes to assigning a specific interpretation to a linguistic string, and how we can model their interactions to make better predictions about human sentence processing.   


Studying LMs' processing of language illusions provides a way to explore whether they can be viewed as cognitive models of human sentence processing. As large language models like ChatGPT improve at generating grammatically appropriate strings, it becomes ever more important to investigate whether they are comparable to human language processing behavior at all \cite[see][for a review]{mahowald_dissociating_2023}. 
From there, we can reason about what characteristics in the training of LMs, the architecture of LMs, and the ``abilities'' of LMs enable them to carry out either literal interpretations and detect the anomaly, or to fall into the illusion rabbit hole. 

\begin{table*}[ht!]
\centering \small
\begin{tabular}{cccccccccc}
\hline
Illusion type & item & \multicolumn{2}{c}{BERT} & \multicolumn{2}{c}{RoBERTa}&\multicolumn{2}{c}{GPT-2}&\multicolumn{2}{c}{GPT-3}\\
\hline
&  & PPL & Surp &PPL & Surp &PPL & Surp &PPL & Surp\\ 
\hline
Comparative & 32 & \cellcolor{LightBlue1} -0.36 & -0.001 & \cellcolor{LightBlue1} -0.56 & \cellcolor{LightBlue1} -0.09 & \cellcolor{LightBlue1} -0.22 & -0.05 & \cellcolor{LightBlue1} -0.30 & \cellcolor{LightBlue1} -0.25\\
\hline
Depth-charge & 32 & \cellcolor{LightBlue1} -0.37 & \cellcolor{LightBlue1} -0.15 & \cellcolor{LightBlue1}-0.61 & \cellcolor{LightBlue1}-0.45 & -0.12 & \cellcolor{LightBlue1} -0.41 & \cellcolor{LightBlue1}-0.37 & \cellcolor{LightBlue1} -0.98 \\
\hline
NPI & 32 & \cellcolor{LightBlue1} -0.26 & \cellcolor{LightBlue1} -2.46 & \cellcolor{LightBlue1} -0.71 & \cellcolor{LightBlue1} -2.60 & \cellcolor{LightBlue1} -0.21 & \cellcolor{LightBlue1} -1.73 & \cellcolor{LightBlue1}-0.29 & \cellcolor{LightBlue1} -2.55\\
\hline
\end{tabular}
\caption{\label{grammaticality_stat_estimate} Estimated coefficients of the main effect (acceptable sentence condition vs. unacceptable condition (reference)) for each statistical model. If LMs rate acceptable sentences as more acceptable, the coefficients for perplexity or surprisal should be significantly negative. Cells color-coded in \hl{blue} represent statistical significance level ($p < .05$) in the expected direction. White cells represent an insignificant main effect. In other words, \hl{blue} cells indicate the statistical model output supports LMs' ability to distinguish sentences based on linguistic acceptability. 
}
\end{table*}

\section{Methods}

\subsection{Models and Measures} \label{sec: model and measure}

We analyzed four models, two masked language models, and two autoregressive models: BERT \citep{devlin_bert_2019}, RoBERTa \citep{liu_roberta_2019}, GPT-2 \citep{radford_language_2019}, GPT-3 \citep{brown_language_2020}. BERT, RoBERTa, and GPT-2 were accessed via HuggingFace \citep{wolfTransformersStateoftheArtNatural2020}, and GPT-3 via OpenAI's API.\footnote{We used `bert-base-cased', `roberta-base', `gpt2', and `text-davinci-003'. Code for replicating the results, statistical tests, and figures can be found at \url{https://github.com/forrestdavis/LanguageIllusions.git} .} We used two measures, sentence level perplexity and surprisal of specific target words. For autoregressive models, the surprisal of a specific word\footnote{For words that are subworded, the joint probability was calculated.} is given by the following equation: 

\begin{equation}
    \textrm{Surp}(\textrm{w}_i) = 
    -\textrm{log Prob(w}_i \vert \textrm{w}_1 ... \textrm{w}_{i-1})
\end{equation}

Perplexity for a sentence of N words is: 

\begin{equation}
    2^{\frac{1}{N} \sum_{i=1}^{N}{\textrm{Surp}(\textrm{w}_i)}}
\end{equation}

For bidirectional models, we calculated the surprisal of a word in a context by using the masking technique in \citet{kauf-ivanova-2023-better}, which corrects for words that are subworded.\footnote{For example, consider the word `souvenir'. This is subworded by BERT into `so', `\#\#uven', and `ir'. Rather than MASK each subpart, one at a time, (e.g. `so' [MASK] `ir'), the right context of the target subword is always masked (e.g. `so' [MASK] [MASK]).} 
Further, we used this masking technique to calculate the pseudo-perplexity of a sentence \citep{salazar-etal-2020-masked}.

\subsection{Evaluation procedure}

We treated LMs as psycholinguistic research subjects to generate both whole-sentence perplexity and surprisals at critical words for carefully controlled minimal pairs for each illusion \cite[following,][]{futrell_neural_2019}. Assuming these two scores are correlated to human acceptability judgments \cite{lau_grammaticality_2017}, we constructed mix-effects linear regression models from the $R$ package $lme4$ to test whether LMs were also sensitive to reported manipulations that affect human judgments. For each scoring metric, we took it as the dependent variable and coded the manipulation condition representing a certain hypothesis into the independent variable. We read the estimated coefficient(s) of the tested condition variable(s) to infer whether LMs show sensitivity to the effect of that condition manipulation on the scoring metric. We evaluated language models in three broad aspects: acceptability differentiation, illusion effect, and sensitivity to manipulations.

\begin{itemize}

    \item \textbf{Acceptability differentiation} We first asked whether language models could distinguish acceptable sentences from unacceptable sentences that humans have no trouble dealing with.\footnote{According to finer-grained linguistic criteria, acceptable sentences are those that are grammatical, plausible, and felicious. Please refer to \citet{tonhauser_empirical_2015} for detailed definitions and review.} Models with relevant knowledge should assign lower perplexity/surprisal to acceptable sentences versus unacceptable ones. 
    
    \item \textbf{Illusion effect} We took the results from the acceptability differentiation task as the foundation to test the illusion sentences. Here, we hypothesized that language models should either (i) align with humans' illusionary judgments, reflected by models' generating a lower perplexity/surprisal for illusion sentences than the unacceptable controls, or (ii) deviate from human behavior and show hints of being a literal processor, reflected by models' generating a higher or similar perplexity/surprisal score compared to the unacceptable condition. If models behave like humans, then we expected (i) to be the models' consistent behavior. If models conform to (ii), we take this as evidence of non-human-like behavior.
    
    \item \textbf{Sensitivity to manipulations} Lastly, we assessed whether language models were sensitive to illusion-specific linguistic manipulations that affect human judgments. A greater degree of sensitivity indicates that the corresponding linguistic knowledge and how the knowledge affects sentence acceptability could be encoded in or learned by LMs. This allowed us to draw a fine-grained comparison between humans and LMs. If language models are insensitive, that indicates a difference between humans and LMs. 
\end{itemize}

\begin{figure*}[ht!]
    \centering
    \includegraphics[width=0.7\textwidth]{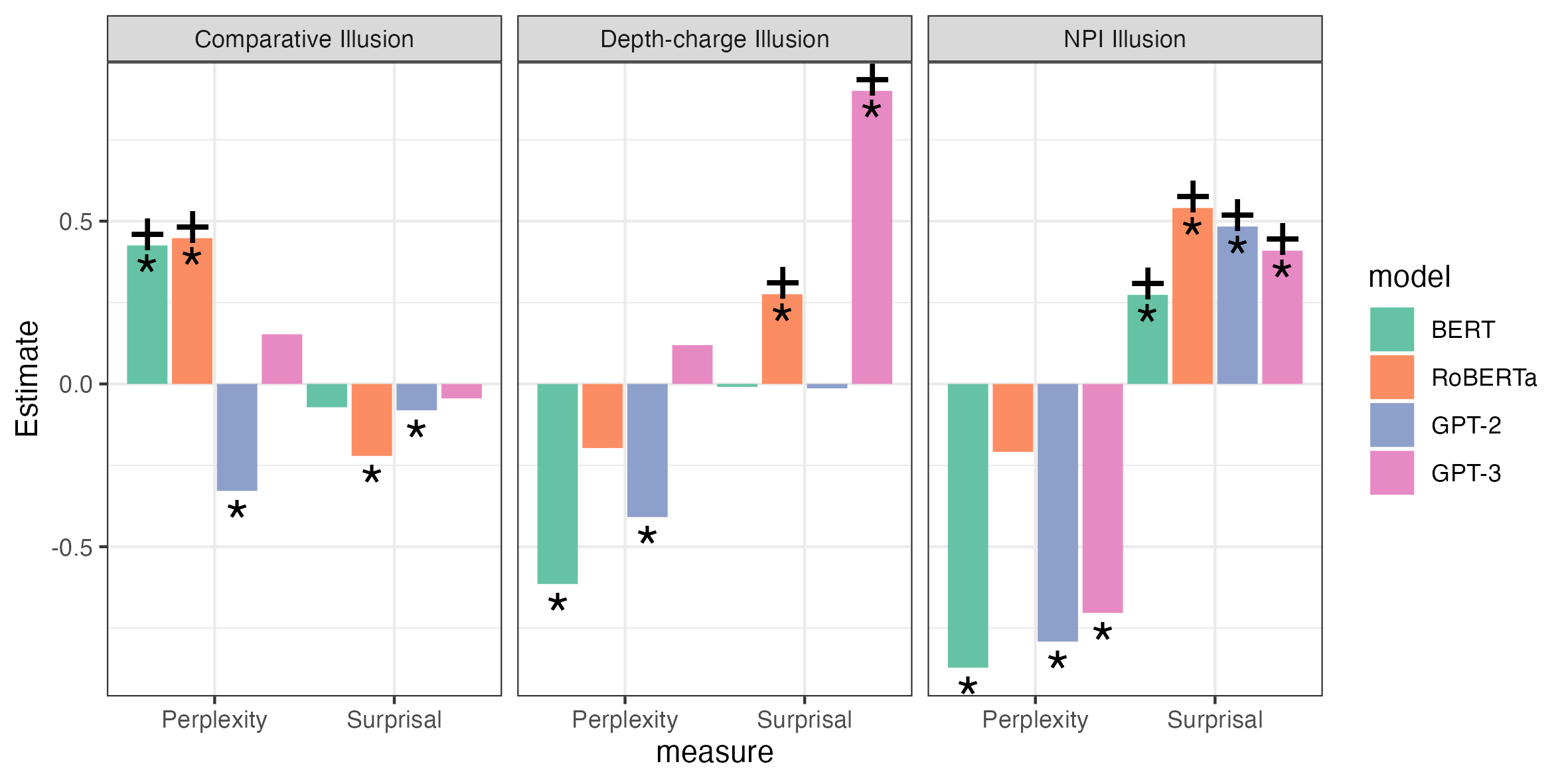}
    \caption{The $y$ axis shows the coefficient estimates which represent the increase in perplexity/surprisal when the sentence is unacceptable compared to the illusion case, crossing three language illusions and four LMs. ``+'' marks a human-like behavior, in this case, an illusion effect where the unacceptable condition receives significantly higher perplexity/surprisal values than the illusion condition. ``*'' means that the estimated coefficient is significant.}
    \label{fig: illusion_effect_estimates}
\end{figure*}

\section{Comparative illusion}\label{section:ci}

A canonical comparative illusion surfaces in sentences like ``More people have been to Russia than I have''. People accept it at first glance but have trouble pinning down the exact meaning \cite{montalbetti_after_1984} one of which could be that the number of the group of people who've been to Russia is greater than the number of ``me''. Potential rational nonliteral inference could be ``people have been to Russia more times than I have'' or ``people have been to Russia but I haven't'' \cite{oconnor_comparative_2015,christensen_dead_2016}. Psycholinguistic research has found that various factors modulate the strength of the illusion, including the repeatability of the event described by the verb phrase, the subject form of the than-clause subject (e.g. ``... than the student has'' vs. ``...I have''), as well as the number of that subject (e.g. ``I have'' vs. ``we have'')\cite{wellwood_anatomy_2018}. There is also a claim arguing that the processing mechanism follows the noisy-channel predictions under an information-theoretic account \cite{zhang2023comparativeillusion}.

We adapted the experimental materials with 32 items from \citet{zhang2023comparativeillusion}.\footnote{See Table \ref{ci_full_paradigm} in the Appendix for the full paradigm.} An example is in \ref{ci_example} where \ref{ci} is the canonical comparative illusion, \ref{ci_pl} is the acceptable control, and \ref{ci_impl} is the unacceptable one.\footnote{The repeatability of the verb phrase is responsible for this contrast, as it is more natural to say ``use Tiktok more often or frequent'' compared with ``install Tiktok more often'' when the action typically takes place once (in a while).}  

\ex. \label{ci_example}
\a.(?) More teenagers have used Tiktok than I \underline{have}. (illusion) \label{ci} 
\b.Many teenagers have used Tiktok more than I \underline{have}. (acceptable) \label{ci_pl}
\c.(\#) Many teenagers have installed Tiktok more than I \underline{have}. (unacceptable) \label{ci_impl}

\subsection{Acceptability differentiation}

We first ensured that LMs distinguish acceptable neighbors \ref{ci_pl} of the illusion sentence from unacceptable ones \ref{ci_impl}. We ran statistical mixed-effects linear regression models on whole-sentence perplexity and the surprisal at the word \textit{have} for the four language models. Either the perplexity or the surprisal was taken as the dependent variable with the condition ``acceptability'' as the fixed effect (reference level = the unacceptable condition, with a nonrepeatable verb phrase vs. the acceptable condition, with a repeatable verb phrase) and the random intercept of each item as the random effect.\footnote{The model syntax in $R$ was \texttt{PPL/SURP $\sim$ acceptability + (1|item).} } 

Table \ref{grammaticality_stat_estimate} shows the estimated coefficient for the main effect of each mixed-effect model for each LM and each illusion phenomenon. A significant negative estimated coefficient suggests that acceptable sentences received lower perplexity/surprisal compared to the unacceptable ones, indicating that LMs distinguish sentences based on acceptability. Except for surprisal values from BERT and GPT-2, the other six statistical models indicate that the LMs capture the acceptability difference of baseline sentences for the comparative illusion.

\subsection{Illusion effect} 

This task investigated whether language models pattern with humans in demonstrating illusion effects with the basic comparative illusion construction. The contrast involves the illusion condition \ref{ci} with existing control conditions (\ref{ci_pl} and \ref{ci_impl}). The standardized metrics of the four LMs are displayed in Figure \ref{fig: illusion_effect} in the Appendix. To evaluate whether LMs capture an illusion effect, we constructed another suite of statistical models across the four LMs and two metrics where the main effect has three levels -- the illusion condition (reference), the acceptable condition, and the unacceptable condition -- and the random effect included a random intercept for items.\footnote{The model syntax in R was \texttt{PPL/SURP $\sim$ condition + (1|item)} where \texttt{condition} had three levels.}

We analyzed the coefficient estimates of the main effect of the unacceptable condition compared with the illusion condition.\footnote{The coefficients for the acceptable condition generate similar conclusions. Further, no illusion sentences were rated better than acceptable ones.} An illusion effect would appear with higher perplexity/surprisal for the unacceptable condition compared to the illusion case. In other words, the estimated coefficients for the unacceptable condition should be significantly positive. 

Figure \ref{fig: illusion_effect_estimates} and Table \ref{illusion_effect} (in Appendix) display the estimated coefficients for the unacceptable condition compared with the illusion condition. For the comparative illusion, only BERT and RoBERTa measured by perplexity show a human-like illusion effect. Other LM-metric combinations indicate that the illusion condition was rated either the same or worse than the unacceptable condition (contrary to humans).

\subsection{Sensitivity to manipulations} \label{sec:ci_linguistic_manipulations}

In this step, we evaluated whether language models were sensitive to sentence manipulations that affect human judgments. Three factors were investigated: (1) than-clause subject structure (pronoun vs. NP), (2) subject number (singular vs. plural), and (3) verb repeatability (repeatable vs. nonrepeatable). For humans, plural than-clause subjects are more acceptable than singular ones only in the NP case. Overall, repeatable verbs are more acceptable than nonrepeatable ones \cite{oconnor_comparative_2015, wellwood_anatomy_2018, zhang2023comparativeillusion}.

\begin{figure}[t!]
    \centering
    \includegraphics[width=\linewidth]{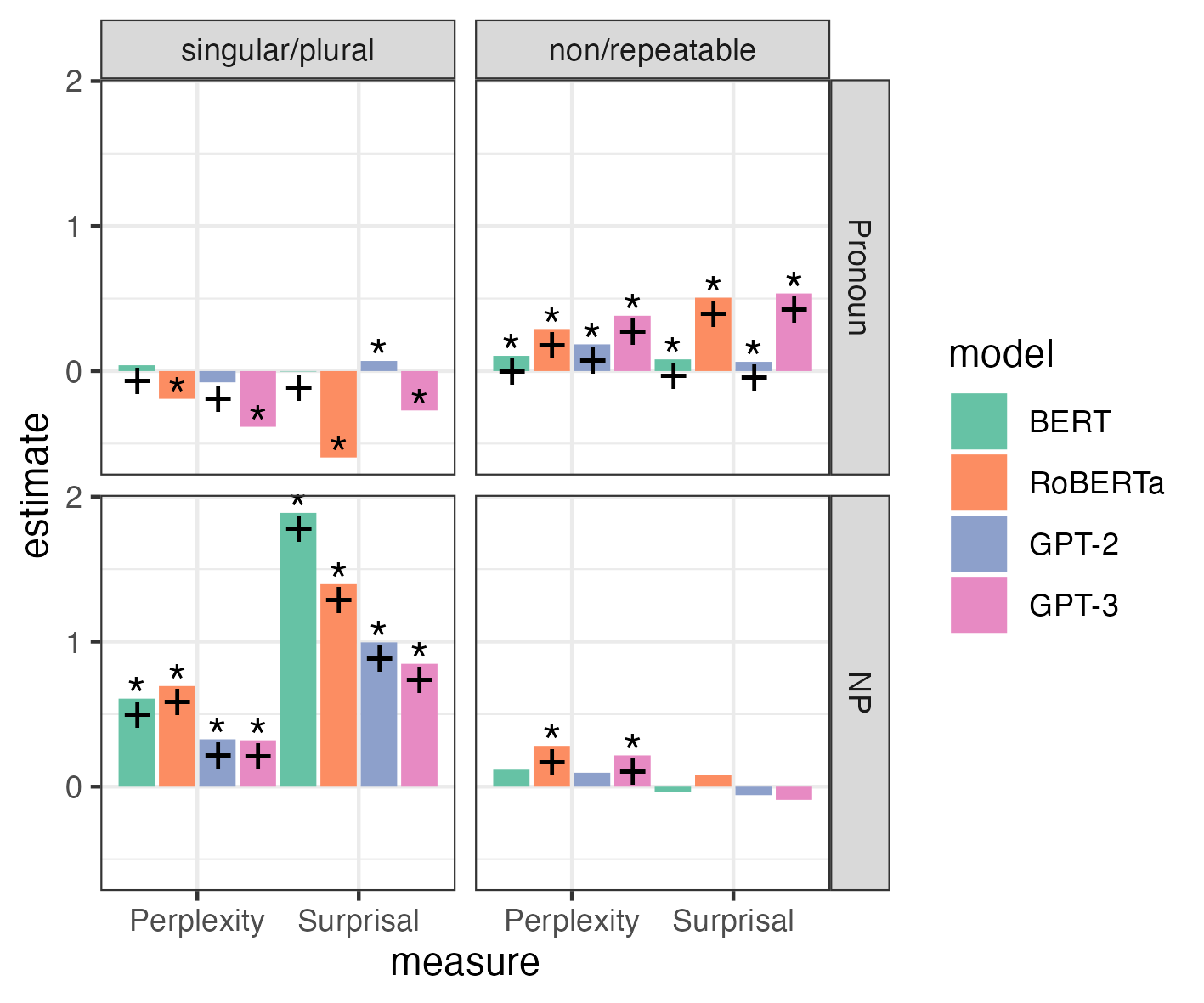}
    \caption{Estimated coefficients for critical linguistic manipulations in \textbf{comparative illusion}. The $y$ axis shows the estimated coefficients for the increase in perplexity/surprisal with respect to singular vs. plural than-clause subjects, or nonrepeatable vs. repeatable verb phrases, respectively. ``*'' means statistically significant contrasts; ``+'' means human-like results.}
    \label{fig: ci_model_estimate}
\end{figure}

Figure \ref{fig: ci_model_estimate} displays the estimated coefficients for the main effects from the statistical models.\footnote{More statistic model information: Iterating over LMs, metrics, and the subject structure (NP vs. pronoun), we initiated statistical models taking both repeatability (reference = repeatable) and subject number (reference = plural) as the main effects with the random effect including a random intercept for the items.} As for the subject number, when the than-clause subject was a pronoun, only BERT and GPT-2 (with perplexity) aligned with human-like behavior: there is no difference between singular and plural than-clause subjects. When it comes to NP subjects, all four LMs with both metrics showed human-like behavior where the singular NP subject was more unacceptable than the plural NP subject. As for repeatability, all four LMs captured this distinction in the pronoun condition but in the NP condition, only RoBERTa and GPT-3 achieved human-like results with perplexity. 

In general, we only found partial overlap between LMs and humans. This indicates that even though LMs show some knowledge of acceptability for comparative structures, they might operate differently from humans when processing more subtle differences. None of the language models fully captured all the manipulations.

\section{Depth-charge illusion} \label{section:dc}
Consider the most famous depth-charge sentence \textit{No head injury is too trivial to be ignored} \cite{wason_verbal_1979}. People overwhelmingly interpret it as meaning ``no matter how trivial head injuries are, we should not ignore them'', while the literal meaning is the opposite as ``we should ignore them''. 

To understand the depth-charge sentence requires knowing meaning composition rules, multiple negation processing \cite{wason_verbal_1979}, adequate world knowledge reasoning \cite{paape_quadruplex_2020}, and the neighboring constructions of \textit{too...to} such as \textit{so...that}, \textit{so...as to} and \textit{enough to...} \cite{zhang2023noisy}. Since existing research already shows that language models are quite limited in processing negation \cite[e.g.][]{DBLP:journals/corr/abs-1911-03343,ettinger_what_2020}, we speculate that LMs might encounter difficulty in the more complicated case of depth-charge sentences.

The evaluation materials were adapted from \citet{zhang2023noisy} with 32 items. An example is \ref{dc_example} where we take the surprisal of the sentence-final word for comparison.

\ex. \label{dc_example}
\a. (?) No head injury is too trivial to be \underline{ignored}. (depth-charge sentence) \label{dc}
\b. Some head injury is too severe to be \underline{ignored}. (plausible, acceptable) \label{dc_pl_control}
\c. (\#) Some head injury is too trivial to be \underline{ignored}. (implausible, unacceptable) \label{dc_impl_control}

\subsection{Acceptability differentiation} 

Utilizing the same methodology as the comparative illusion, we found, as depicted in Table \ref{grammaticality_stat_estimate}, that all combinations of LMs and metrics, except GPT-2 (perplexity), captured the acceptability difference between (\ref{dc_pl_control}) and (\ref{dc_impl_control}) with a significantly lower perplexity/surprisal for the acceptable sentences like \ref{dc_pl_control}.

\subsection{Illusion effect} 

Next, we studied if LMs ``experience'' the illusion effect by assigning lower perplexity/surprisal scores to the depth-charge sentence \ref{dc} compared to the unacceptable one \ref{dc_impl_control}.

Our statistical results show, in Figure \ref{fig: illusion_effect_estimates} and Table \ref{illusion_effect} (Appendix), that only RoBERTa and GPT-3 demonstrated an illusion effect (for surprisal) by assigning a significantly higher score to the unacceptable control sentences. This means that it is not easy to ``trick'' LMs with the depth-charge illusion. Similar results have led concurrent work to suggest that LMs are better at deriving the literal meaning of a sentence, which is in sharp contrast with the overwhelming illusion effect from humans \cite[][a.o.]{paape_when_2023}.
 
\subsection{Sensitivity to manipulations} \label{sec:dc_linguistic_manipulations}
 
This task tested LMs' sensitivity to the plausibility contrast of three near-neighbor pairs of the depth-charge sentence. These pairs differ by the degree quantifier construction (\textit{too...to} vs. \textit{so...as to} vs. \textit{too...to not}).\footnote{The full suite of paradigms is shown in Table \ref{dc_full_paradigm} in the Appendix.} Competent language models should differentiate plausible sentences from implausible ones.

\begin{figure}[t!]
    \centering
    \includegraphics[width=\linewidth]{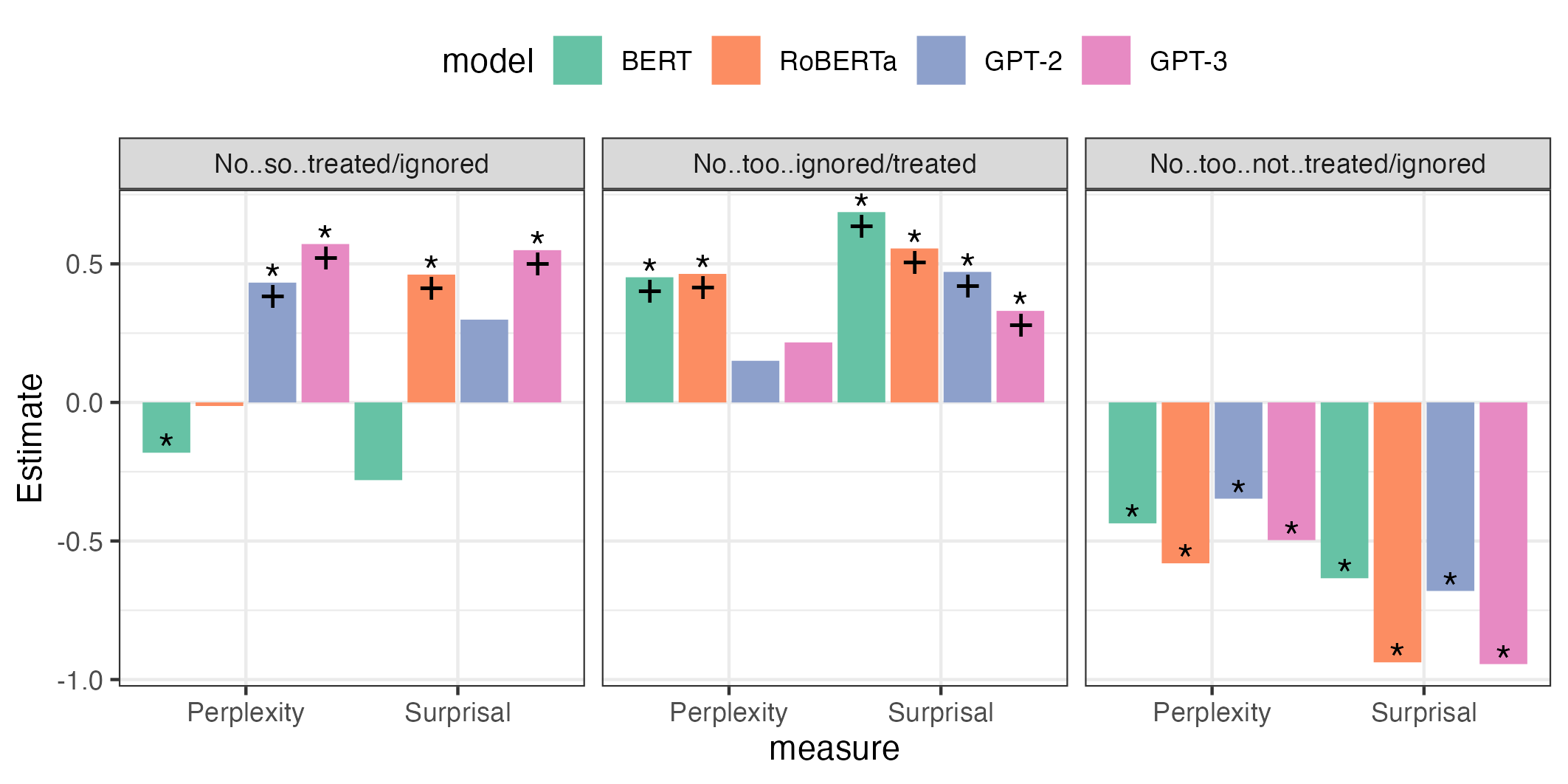}
    \caption{Estimated coefficients for the plausibility contrast (reference = plausible) in \textbf{depth-charge illusions}. The $y$ axis shows the increase in perplexity/surprisal when the sentence is implausible vs. plausible. ``*'' means statistically significant contrasts; ``+'' means human-like behavior. While we see differences among LMs and metrics in the ``no...so...as to'' and the ``no...too...to'' conditions, the condition of ``no...too...to not'' yielded completely opposite results to humans.}
    \label{fig: dc_model_estimate}
\end{figure}

Figure \ref{fig: dc_model_estimate} displays estimated coefficients of statistical models' main effect. We expect implausible sentences to receive higher perplexities/surprisals when the illusion occurs.\footnote{Iterating over sentence pairs, LMs, and metrics, we ran mixed-effects linear regression models on scores over the plausibility contrast (reference = plausible).} We find that LMs captured some of the distinctions in the \textit{too...to} condition and the \textit{so...as to} condition. However, implausible sentences with \textit{too...to not} were rated as more acceptable than their plausible counterparts, which flouts what linguistic rules predict.\footnote{The sentence \textit{No head injury is too trivial to not be ignored} should be plausible because compositionally, ``too trivial to not be ignored'' means ``too trivial to be treated'' which yields a plausible sentence given the sentential negation.} The fact that ``No head injury is too trivial to be treated'' and ``No head injury is too trivial to not be ignored'' generate opposite results while having the same meaning suggests LMs still struggled with negation, antonyms, and meaning composition \cite{kim-2020-cogs,she2023scone,truong2023language}.

\section{NPI illusion} \label{section:npi}

Negative polarity items and their licensing conditions have been investigated in prior work with language models. For a canonical NPI (e.g. \textit{ever}, \textit{any}) to be acceptable, it has to be in the scope of negation.\footnote{The licensing conditions of negative polarity items are far more than in the scope of negation. We focus on the classic licensing condition and refer to \citet{giannakidou2019negative} for a review.} Existing computational research has shown that the syntactic dependency between the licensor and the NPI is captured by language models \cite{jumelet_language_2018,jumelet_language_2021,shin_investigating_2023} but with more difficulty as compared to subject-verb agreement or other syntactic dependencies \cite{marvin-linzen-2018-targeted,warstadt_investigating_2019,warstadt_blimp_2020}. In this task, we expanded the suite of LMs and metrics and explored sensitivities to four types of licensors. 

Our materials came from \citet{orth_negative_2021} with 32 items. The essential triad is \ref{npi_example} where the illusion condition has the NPI \textit{ever} not in the scope of the negation word \textit{no}.

\ex. \label{npi_example}
\a. (?) The hunter who no villager believed to be trustworthy will \underline{ever} shoot a bear. (NPI illusion) \label{npi_ill}
\b. No hunter who the villager believed to be trustworthy will \underline{ever} shoot a bear. (Matrix No, acceptable) \label{npi_good}
\c. (*) The hunter who the villager believed to be trustworthy will \underline{ever} shoot a bear. (Licensor Absent, unacceptable) \label{npi_bad}

\subsection{Acceptability differentiation} 

Table \ref{grammaticality_stat_estimate} shows that all the four LMs could capture the acceptability difference of control sentences \ref{npi_good} and \ref{npi_bad} (with both metrics).

\subsection{Illusion effect} \label{sec: npi_illusion_effect}

Figure \ref{fig: illusion_effect_estimates} and Table \ref{illusion_effect} show that only in the case of surprisal did we see an illusion effect where the unacceptable sentences (e.g. \ref{npi_bad}) received significantly higher surprisals than the illusion sentence (e.g., \ref{npi_ill}). This finding replicates \citet{shin_investigating_2023} in that, for the illusion condition (\ref{npi_ill}) where \textit{no} linearly precedes \textit{ever} but is in an unlicensing position, \textit{ever} incurs higher surprisal. It is interesting to see the sharp discrepancy between surprisal and perplexity, which we leave to Section \ref{dis_ppl_surp} for discussion.

\begin{figure}[t!]
    \centering
    \includegraphics[width=\linewidth]{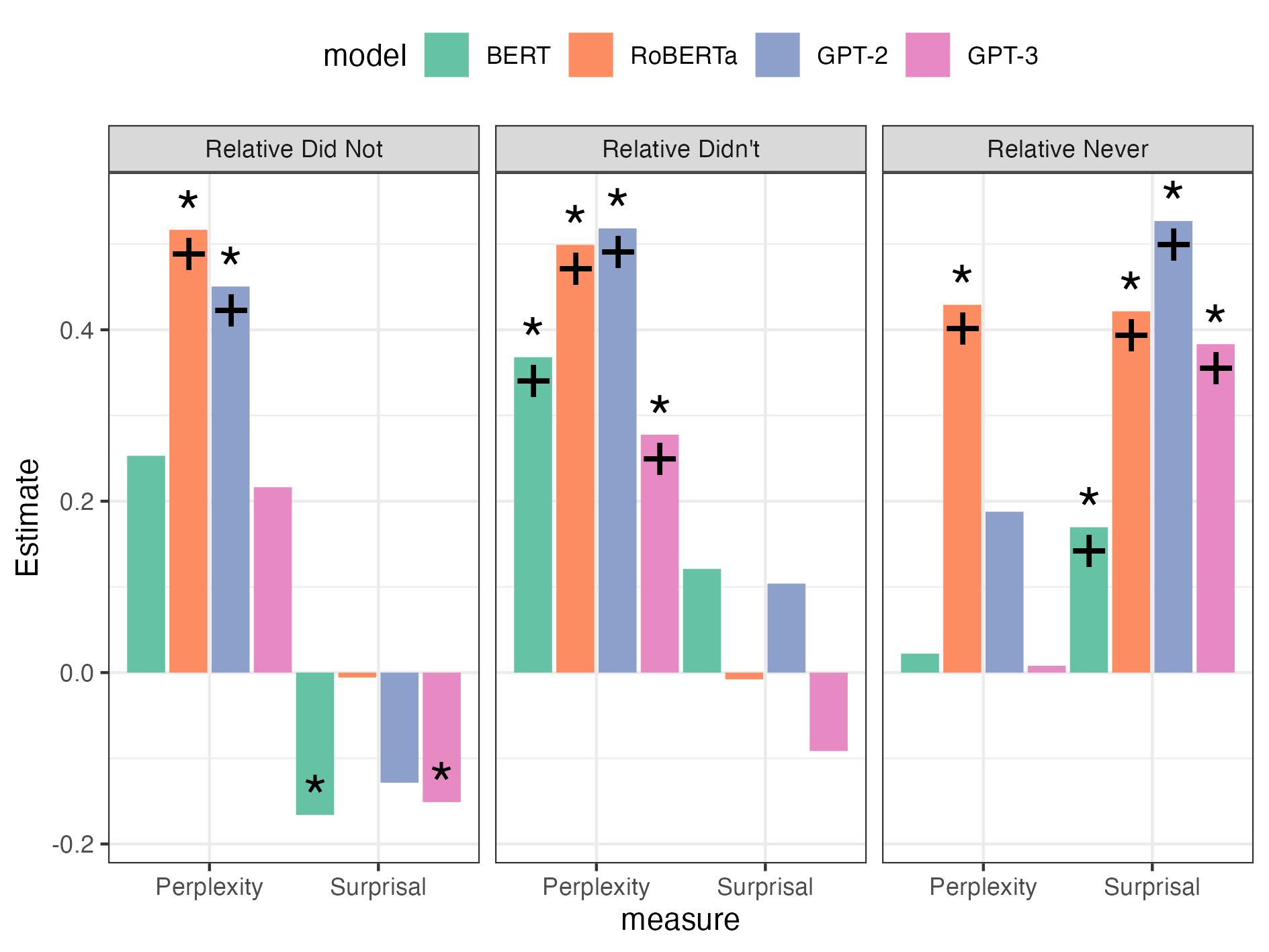}
    \caption{Estimated coefficients for the illusion effect (unacceptable vs. illusion = reference) in \textbf{NPI illusions}. The $y$ axis shows the increase in perplexity/surprisal when the sentence is ungrammatical vs. is in the illusion condition. ``+'' marks an illusion effect while none of the three licensors should trigger an illusion effect according to human behavior; ``*'' means a significant contrast.}
    \label{fig: npi_model_estimate}
\end{figure}

\subsection{Sensitivity to variations} 

The linguistic manipulations we explored concern the illusion effect in the illusion condition with different NPI licensors. Among the ones we tested, \textit{didn't}, \textit{did not}, and \textit{never},\footnote{Please refer to Table \ref{npi_full_paradigm} for the full experimental conditions.} human research shows that none of these triggers illusion effects (\citealp{orth_negative_2021}; \citealp[cf.][]{vasishth_processing_2008}).

Iterating over licensors, LMs, and metrics, we ran statistical models with the same structure in Section \ref{sec: npi_illusion_effect}. We plotted the estimated coefficients of the unacceptable main effect in Figure \ref{fig: npi_model_estimate} and predicted that a significantly positive coefficient indicates an illusion effect. Contrary to human-like behavior, for all three licensors there were some LM-metric combinations that indicate an illusion effect: for the licensor \textit{did not}, RoBERTa (perplexity) and GPT-2 (perplexity) show an illusion effect; for \textit{didn't}, all four LMs with perplexity show an illusion effect; for \textit{never}, all four LMs with surprisal, plus RoBERTa with perplexity, show an illusion effect. This pattern shows that with NPI illusions, LMs are more easily tricked than humans. 

\section{Discussion}

\subsection{Illusion effect}

Successful language processing requires a dynamic integration of lexical knowledge, grammatical knowledge, logical reasoning, and world knowledge, among other cognitive abilities and sources of knowledge. An illusion effect in humans where unacceptable sentences receive unexpectedly high acceptability presents a unique case where the comprehender might prioritize different processing mechanisms or linguistic constraints for meaning inference over those employed for common processing. Studying how language models process language illusions helps us understand (1) from a superficial level, whether LMs appear to be human-like -- circumventing some grammatical facts and reaching a good-enough sentence representation, and (2) from a deeper level, whether LMs employ the same set of resources and abilities to process a sentence (i.e.\ whether they can serve as cognitive models).

\begin{figure*}[t!]
    \centering
    \includegraphics[width=0.8\linewidth]{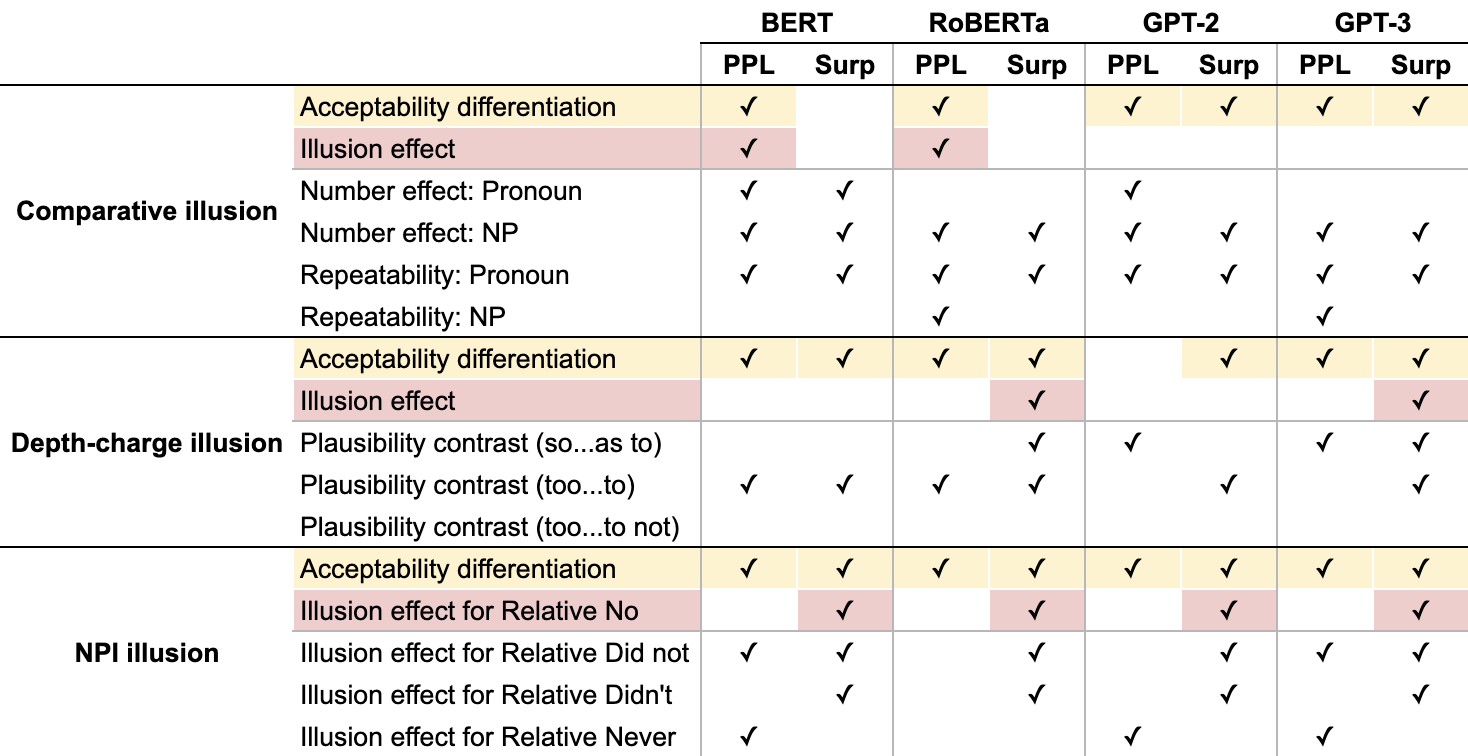}
    \caption{Language models' performance on all three illusions. \checkmark means LMs show human-like behavior.}
    \label{fig: table_of_summ}
\end{figure*}

In this research, we aim for the first level of understanding. By studying four language models' acceptability judgments of three language illusions, we found that LMs were good at the basic acceptability differentiation task and yet no LMs showed consistent human-like illusion effects among three illusion phenomena by any metric (Figure \ref{fig: table_of_summ}). We conclude from this result that LMs might not be a good cognitive model of human language processing. With this said, we do observe a divergence between the comparative/depth-charge illusion and the NPI illusion -- it seems more likely for LMs to be tricked by the NPI illusion compared to the former two. Since the NPI illusion is more relevant to the hierarchical structure of language whereas both the comparative illusion and depth-charge illusion emphasize semantic nuances, we tentatively conclude that LMs are more easily tricked by syntactic illusion rather than semantic illusions.

\subsection{Human-like behaviors \& Potential processing mechanisms}

For both the comparative illusion and depth-charge illusion, the illusion effect test did not show human-like behavior. This could either mean that LMs strictly abide by linguistic rules to compose the language literally or that LMs have trouble understanding this complicated set of sentences overall. For the comparative illusion, the sensitivity task (Section \ref{sec:ci_linguistic_manipulations}) suggests that they might have some capacity to process comparative structures. For the depth-charge illusion, that LMs seem to have trouble understanding the literal contrast between plausible/implausible pairs (Section \ref{sec:dc_linguistic_manipulations}) suggests sentences involving multiple negations could pose a challenge to LMs. The two cases indicate we still need to develop more robust evaluations to gauge LMs' semantic capabilities in various semantic domains.

For the NPI illusion, the interpretation could be more complicated. On one hand, the illusion test for the licensor \textit{no} yields human-like results (with surprisal) but other licensors also elicit non-human-like illusion effect \cite[cf.][]{orth_negative_2021}. On the other hand, the discrepancy between sentence perplexity and surprisal makes it difficult to conclude to what degree LMs and humans overlap \cite[cf.][]{shin_investigating_2023}.

Ultimately, we want to address whether LMs are like humans that utilize not only grammatical rules but also contexts, frequencies, and semantic priors to rationally process language, or LMs are like grammarians that interpret string inputs in a strict compositional manner. Our investigation does not yield consistent results given the three language illusions but the behavioral inconsistency suggests that language models are far from being a cognitive model of human language.

\subsection{Language models' performance in general}

All four language models performed on par with each other. If we tallied the number of tests where LMs reported expected results from Figure \ref{fig: table_of_summ} and averaged between perplexity and surprisal, we have a ranking order from RoBERTa (N=10) and GPT-3 (N=9), to BERT (N=8.5) and GPT-2 (N=8). The successors of both the masked language model and the autoregressive model perform better than their predecessors.

\subsection{Perplexity \& Surprisal}\label{dis_ppl_surp}




It is surprising to see that the two widely used probability-based metrics can generate different results for a given hypothesis and a given language model. Future work should (i) investigate both mathematically and practically why the difference could occur and (ii) check if better definitions for the critical regions exist to capture surprisals. Future evaluation work that utilizes one metric should be mindful of the intrinsic limitations of that metric.

\subsection{Limitations}

Considering the research methodology, acceptability judgment tasks (even with carefully controlled minimal pairs) are indirect measures of language comprehension and it is hard to infer the exact interpretation based on probability-based measures. Further studies should work on direct comprehension measures (e.g. generating paraphrases) that reveal LMs' hidden knowledge.

\section{Conclusion}

We tested four language models' ability to process three language illusions and asked (1) whether they judge unacceptable illusion sentences to be more acceptable as humans (termed an illusion effect) and (2) whether they are sensitive to linguistic manipulations that modulate human judgments. Our results are based on whole-sentence perplexity and critical word surprisal. We show that none of the LMs demonstrated consistent illusion effects or exhibited overall human-like judgment behaviors. We conclude that given the case of language illusions, language models neither behave like humans with full sets of cognitive abilities and error-prone behavior nor possess the necessary linguistic knowledge for error-free, literal sentence processing. Language models cannot be viewed as cognitive models of language processing, which makes understanding them even more intriguing.



\section*{Acknowledgements}
We thank the three anonymous reviewers for their helpful feedback. We thank Ankana Saha, Carina Kauf and Hayley Ross for the helpful discussion about the project. 

\input

\bibliographystyle{acl_natbib}

\appendix \label{sec:appendix}

\begin{table*}
    \centering
    \begin{tabular}{ccccccccc}
    \hline
        Illusion type & \multicolumn{2}{c}{BERT} & \multicolumn{2}{c}{RoBERTa}& \multicolumn{2}{c}{GPT-2} & \multicolumn{2}{c}{GPT-3} \\ \hline
        ~ & PPL & Surp & PPL & Surp & PPL & Surp & PPL & Surp  \\ \hline
        Comparative & \cellcolor{DarkSeaGreen2}\textbf{0.43} & \cellcolor{PeachPuff1} -0.07 & \cellcolor{DarkSeaGreen2}\textbf{0.45} & \cellcolor{LightSalmon1}\textbf{-0.22} & \cellcolor{LightSalmon1}\textbf{-0.33} & \cellcolor{LightSalmon1}\textbf{-0.08} & \cellcolor{PeachPuff1}0.15 & \cellcolor{PeachPuff1}-0.04  \\ \hline
        Depth-charge & \cellcolor{LightSalmon1}\textbf{-0.61} & \cellcolor{PeachPuff1}-0.01 & \cellcolor{PeachPuff1}-0.20 & \cellcolor{DarkSeaGreen2}\textbf{0.28} & \cellcolor{LightSalmon1}\textbf{-0.41} & \cellcolor{PeachPuff1}-0.01 & \cellcolor{PeachPuff1}0.12 & \cellcolor{DarkSeaGreen2}\textbf{0.90}  \\ \hline
        NPI& \cellcolor{LightSalmon1}\textbf{-0.87} & \cellcolor{DarkSeaGreen2}\textbf{0.27} & \cellcolor{PeachPuff1}-0.21 & \cellcolor{DarkSeaGreen2}\textbf{0.54} & \cellcolor{LightSalmon1}\textbf{-0.79} & \cellcolor{DarkSeaGreen2}\textbf{0.48} & \cellcolor{LightSalmon1}\textbf{-0.70} & \cellcolor{DarkSeaGreen2}\textbf{0.41} \\ \hline
        &&&&&&&&\\
        \cellcolor{DarkSeaGreen2}& \multicolumn{8}{l}{Illusion sentences are more acceptable than unacceptable sentences.}\\
        \cellcolor{LightSalmon1}& \multicolumn{8}{l}{The unacceptable sentences are more acceptable than illusion sentences.}\\
        \cellcolor{PeachPuff1}&\multicolumn{8}{l}{No significant difference between the two conditions.}\\
        
    \end{tabular}
    \caption{\label{illusion_effect} Estimates of the main effect (unacceptable sentences vs. illusion sentences) for each statistical model. Positive estimates mean larger perplexity or word surprisals for the unacceptable condition which indicates an illusion effect. Negative estimates mean the unacceptable condition is more acceptable than the illusion condition, which is opposite to the prediction. Bolded estimates represent statistical significance ($p<.05$). We mark the cell in green if there is an illusion effect; in orange for no illusion effect.
}
\end{table*}

\begin{table*}
\centering
\begin{tabular}{lll}
\multicolumn{3}{c}{\textbf{COMPARATIVE ILLUSION}} \\
\hline
\textbf{Number} & \textbf{VP} & \textbf{Examples} \\ 
\hline
\multicolumn{3}{l}{When the \textit{than}-clause subject is \textbf{noun phrase}:} \\
\hline
Singular & Repeatable & More students \textit{have been to Russia} than \textbf{the teacher }has. \\
\hline
Singular & Non-repeatable & More students \textit{have escaped from Russia} than \textbf{the teacher} has.\\ 
\hline
Plural & Repeatable & More students \textit{have been to Russia} than \textbf{the teachers} have. \\
\hline
Plural & Non-repeatable & More students \textit{have escaped from Russia} than \textbf{the teachers} have.\\ 
\hline
Control & Repeatable & More students \textit{have been to Russia} than \textbf{teachers} have. (Good) \\
\hline
Control & Non-repeatable & More students \textit{have escaped from Russia} than \textbf{teachers} have. (Good)\\ 
\hline
\multicolumn{3}{l}{When the \textit{than}-clause subject is \textbf{pronoun}:} \\
\hline
Singular & Repeatable & More students \textit{have been to Russia} than \textbf{I} have. \\
\hline
Singular & Non-repeatable & More students \textit{have escaped from Russia} than \textbf{I} have.\\ 
\hline
Plural & Repeatable & More students \textit{have been to Russia} than \textbf{we} have. \\
\hline
Plural & Non-repeatable & More students \textit{have escaped from Russia} than \textbf{we} have.\\ 
\hline
Control & Repeatable & Many students \textit{have been to Russia} more than \textbf{I} have. (Good) \\
\hline
Control & Non-repeatable & Many students \textit{have escaped from Russia} more than \textbf{I} have. (Bad)\\ 
\hline
\end{tabular}
\caption{\label{ci_full_paradigm}
Full manipulation for the Comparative illusion
}
\end{table*}

\begin{table*}
\centering
\begin{tabular}{lll}
\multicolumn{3}{c}{\textbf{DEPTH CHARGE ILLUSION}} \\
\hline
\multicolumn{2}{c}{\textbf{Conditions}} & \textbf{Examples} \\
\hline
\multicolumn{2}{l}{Canonical depth-charge} & No head injury is too trivial to be ignored.\\
\hline
\multicolumn{2}{l}{Plausible control} & Some head injury is too severe to be ignored.\\
\hline
\multicolumn{2}{l}{Implausible control} & Some head injury is too trivial to be ignored.\\ 
\hline
too...to & plausible & No head injury is too trivial to be treated. \\
\hline
too...to & implausible & No head injury is too trivial to be ignored. \\
\hline
too...to not & plausible & No head injury is too trivial to not be ignored. \\
\hline
too...to not & implausible & No head injury is too trivial to not be treated. \\
\hline
so...as to & plausible & No head injury is so trivial as to be ignored. \\
\hline
so...as to & implausible & No head injury is so trivial as to be treated. \\
\hline
\end{tabular}
\caption{\label{dc_full_paradigm}
Full manipulation for the Depth-charge illusion
}
\end{table*}

\begin{table*}
\centering
\begin{tabular}{ll}
\multicolumn{2}{c}{\textbf{NPI ILLUSION}} \\
\hline
\textbf{Conditions} & \textbf{Examples} \\
\hline
Matrix No & \textbf{No} hunter who the villager believed to be trustworthy will \underline{ever} shoot a bear.\\
\hline
Licensor Absent & The hunter who the villager believed to be trustworthy will \underline{ever} shoot a bear.\\
\hline
Relative No & The hunter who \textbf{no} villager believed to be trustworthy will \underline{ever} shoot a bear. \\
\hline
Relative Didn't & The hunter who \textbf{didn't} believe the villager to be trustworthy will \underline{ever} shoot a bear. \\
\hline
Relative Did not & The hunter who \textbf{did not} believe the villager to be trustworthy will \underline{ever} shoot a bear. \\
\hline
Relative Never & The hunter who \textbf{never} believed the villager to be trustworthy will \underline{ever} shoot a bear. \\
\hline
\end{tabular}
\caption{\label{npi_full_paradigm}
Full manipulation for the NPI illusion}
\end{table*}

\begin{figure*}
    \centering
    \includegraphics[width=\textwidth]{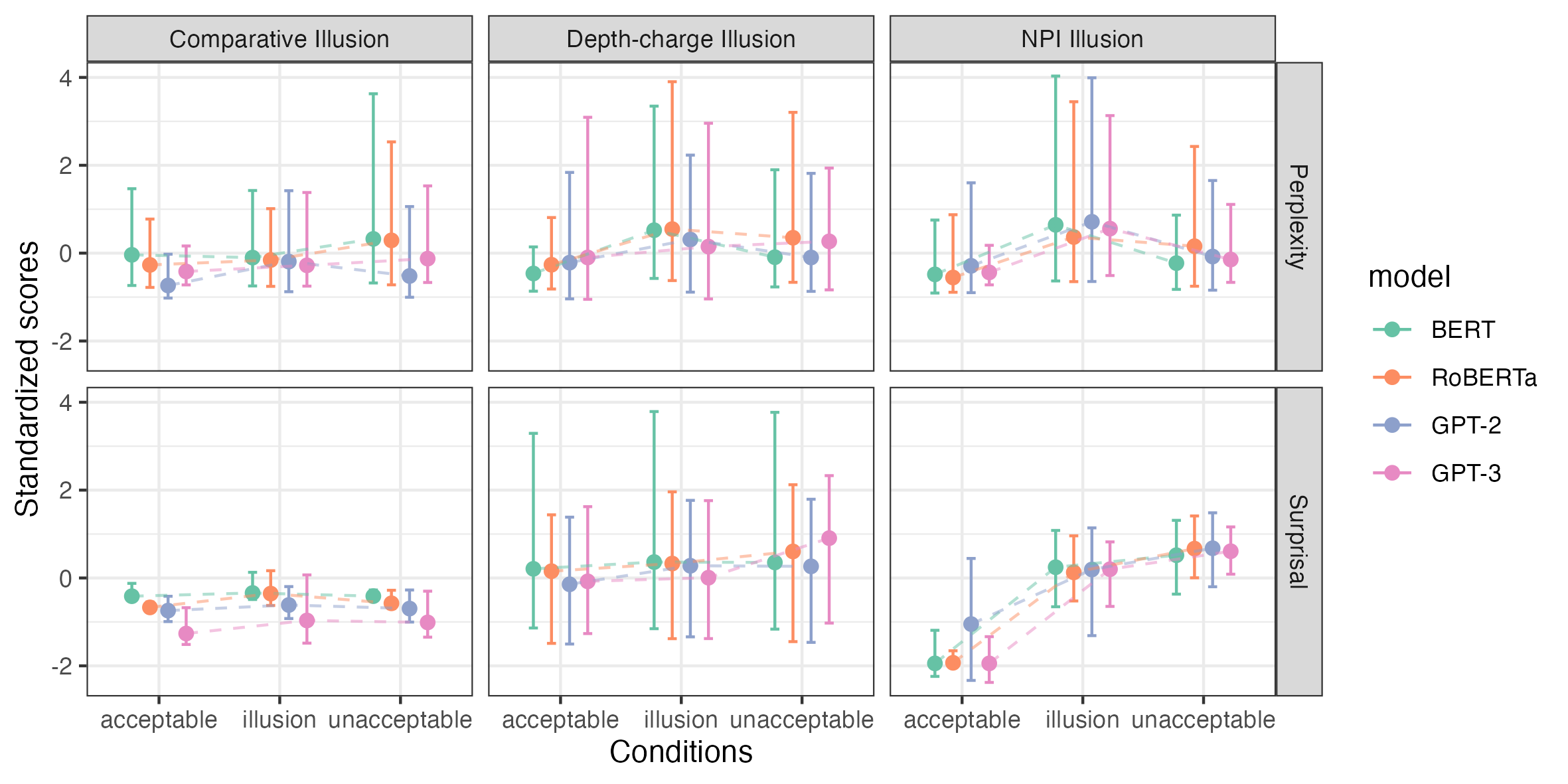}
    \caption{Standardized scores of the Perplexity \& Surprisal for sentences in three conditions crossing LMs and language illusion types. If the illusion effect appears, the illusion condition should be rated more acceptable (thus lower in the graph) than the unacceptable condition and therefore has lower perplexity/surprisal. (Error bars are 95\% bootstrapped confidence intervals).}
    \label{fig: illusion_effect}
\end{figure*}

\end{document}